\documentclass{article}

\usepackage[final, nonatbib]{neurips_2024}
\usepackage[utf8]{inputenc}
\usepackage[T1]{fontenc}
\usepackage{hyperref}
\usepackage{graphicx}
\usepackage{booktabs}
\usepackage{amsfonts}
\usepackage{nicefrac}
\usepackage{microtype}
\usepackage{amsmath}
\usepackage[table,xcdraw]{xcolor}

\title{Setting Standards in Turkish NLP: TR-MMLU for Large Language Model Evaluation}

\author{
M. Ali Bayram$^1$, Ali Arda Fincan$^2$, Ahmet Semih Gümüş$^2$, \\
Banu Diri$^1$, Savaş Yıldırım$^3$, Öner Aytaş$^4$ \\[0.2cm]
$^1$Yıldız Technical University, $^2$Yeditepe University, \\ 
$^3$İstanbul Bilgi University, $^4$Işık University \\[0.2cm]
\texttt{malibayram20@gmail.com}
}

\begin{document}
\maketitle

\begin{abstract}
Language models have made remarkable advancements in understanding and generating human language, achieving notable success across a wide array of applications. However, evaluating these models remains a significant challenge, particularly for resource-limited languages such as Turkish. To address this gap, we introduce the Turkish MMLU (TR-MMLU) benchmark, a comprehensive evaluation framework designed to assess the linguistic and conceptual capabilities of large language models (LLMs) in Turkish. TR-MMLU is constructed from a carefully curated dataset comprising 6,200 multiple-choice questions across 62 sections, selected from a pool of 280,000 questions spanning 67 disciplines and over 800 topics within the Turkish education system. This benchmark provides a transparent, reproducible, and culturally relevant tool for evaluating model performance. It serves as a standard framework for Turkish NLP research, enabling detailed analyses of LLMs’ capabilities in processing Turkish text and fostering the development of more robust and accurate language models. In this study, we evaluate state-of-the-art LLMs on TR-MMLU, providing insights into their strengths and limitations for Turkish-specific tasks. Our findings reveal critical challenges, such as the impact of tokenization and fine-tuning strategies, and highlight areas for improvement in model design. By setting a new standard for evaluating Turkish language models, TR-MMLU aims to inspire future innovations and support the advancement of Turkish NLP research.


\end{abstract}

\section{Introduction}

Advancements in artificial intelligence (AI) and natural language processing (NLP) have revolutionized the field of computational linguistics, particularly with the advent of large language models (LLMs) such as GPT-4, BERT, and Llama. These models, trained on massive datasets, have demonstrated remarkable capabilities in a diverse range of applications, including machine translation, question answering, content generation, and even code synthesis. Their integration into real-world systems has transformed industries, driving innovation and efficiency. However, evaluating the true capabilities of these models remains a persistent challenge, particularly for resource-limited languages like Turkish, where performance often falls short of human-level comprehension \cite{hendrycks2021}.

Existing benchmarks for assessing LLMs predominantly focus on widely spoken languages like English. While these benchmarks, such as MMLU and SuperGLUE, have provided comprehensive evaluation frameworks for high-resource languages, they often overlook the unique linguistic complexities of morphologically rich and agglutinative languages like Turkish. Turkish, with its intricate morphology and syntactic structure, poses distinct challenges for NLP. For instance, a single Turkish verb root can generate numerous word forms through suffixation, encoding tense, person, and mood in a compact yet complex manner. This variability complicates tokenization and semantic parsing, requiring models to grasp both surface patterns and deep linguistic structures.

To effectively evaluate Turkish NLP models, it is essential to account for these unique features. Questions derived from Turkey’s education system offer an ideal basis for such assessments. The education system’s curriculum, with its wide-ranging topics and rich linguistic content, provides a robust framework for evaluating the depth of knowledge and linguistic capabilities of LLMs. These questions test not only factual recall but also conceptual understanding, logical reasoning, and cultural context, making them invaluable for benchmarking Turkish language models.

Two core dimensions underpin the evaluation of LLMs: \textit{instruction following} and \textit{knowledge assessment}. Instruction following assesses a model's ability to execute specific commands, focusing on adherence to predefined instructions and task completion. In contrast, knowledge assessment examines a model's comprehension and understanding, evaluating the breadth and depth of its knowledge base. While instruction-following tasks are important for practical applications, knowledge assessment provides a more comprehensive evaluation of a model's linguistic and conceptual capabilities. For Turkish, where linguistic complexity often obscures semantic nuances, knowledge assessment is particularly critical for evaluating model performance.

This study emphasizes knowledge assessment as the primary metric for evaluating Turkish LLMs. By leveraging questions from standardized Turkish exams such as the University Entrance Examination and the Open Education Faculty (AUZEF) exams, we aim to comprehensively analyze model performance across a diverse set of topics. These exams, designed to test student proficiency across disciplines, provide a rigorous and culturally relevant benchmark for assessing Turkish NLP models. 

The objectives of this study are threefold:
1. To introduce the Turkish MMLU (TR-MMLU) benchmark, a comprehensive evaluation framework tailored for Turkish.
2. To analyze the performance of state-of-the-art LLMs on TR-MMLU, highlighting their strengths and limitations in Turkish-specific tasks.
3. To identify critical challenges and opportunities for improving Turkish NLP through tailored approaches such as advanced tokenization and fine-tuning strategies.

In summary, this study seeks to advance Turkish NLP by addressing critical gaps in model evaluation. By leveraging questions grounded in Turkey’s education system, TR-MMLU provides a robust framework for assessing the linguistic and conceptual capabilities of Turkish LLMs. This benchmark not only sets a new standard for evaluating Turkish language models but also paves the way for future innovations in resource-limited language processing.
\section{Related Work}

The rapid development of large language models (LLMs) has been accompanied by significant efforts to evaluate their performance across various linguistic tasks. Benchmarks and datasets such as MMLU, SuperGLUE, and SQuAD have been instrumental in assessing comprehension, knowledge depth, and generative capabilities, primarily in widely spoken languages like English. However, resource-limited languages, including Turkish, pose unique challenges due to their morphological richness and syntactic complexity. In response, several benchmarks have been developed specifically for Turkish, aiming to address these limitations and provide meaningful evaluations of LLMs.

One of the most comprehensive efforts in Turkish NLP evaluation is \textbf{TurkishMMLU}. This benchmark was developed to meet the need for a structured evaluation framework tailored to Turkish LLMs. The dataset includes questions derived from Turkey’s national curriculum, covering a diverse range of subjects designed to assess linguistic, cultural, and conceptual comprehension. Each question is assigned a difficulty rating based on real-world educational performance, enabling nuanced assessments of model capabilities. TurkishMMLU has been applied to over 20 models, including open-source multilingual LLMs like Llama and MT5, as well as Turkish-adapted models. The benchmark supports various evaluation paradigms, such as zero-shot, few-shot, and chain-of-thought reasoning, making it a foundational resource for Turkish NLP research \cite{yuksel2024}.

Additional Turkish-specific benchmarks include the \textbf{OpenLLMTurkishLeaderboard} and its updated version, \textbf{OpenLLMTurkishLeaderboard\_v0.2}. These benchmarks rely on translated datasets from widely used English benchmarks such as MMLU, AI2\_ARC, and GSM8K. While these resources provide insights into model performance across a range of tasks, translation-based benchmarks often fail to capture the linguistic and cultural nuances unique to Turkish. Consequently, models evaluated on these datasets may exhibit performance inconsistencies when applied to native Turkish tasks. Nevertheless, studies leveraging these benchmarks have demonstrated the effectiveness of fine-tuning multilingual models with Turkish-specific data, highlighting the importance of language-specific training to improve performance \cite{openllm2024, dogan2024}.

The \textbf{THQuAD} dataset, focusing on historical Turkish texts, evaluates reading comprehension tasks using models such as BERT and ELECTRA. This dataset underscores the challenges inherent to Turkish NLP, including context-based understanding and the interpretation of nuanced, historically rooted language. Although THQuAD provides valuable insights into specific domains, its narrow focus on historical texts limits its applicability to broader Turkish NLP tasks \cite{soygazi2021}.

In addition to these structured benchmarks, other studies have explored Turkish NLP tasks such as sentiment analysis, question answering, and language-specific embeddings. Sentiment analysis research, in particular, has highlighted the challenges of processing Turkish social media content, which is often informal and laden with linguistic variations. These efforts underscore the critical need for culturally aligned, Turkish-specific datasets to bridge the performance gap between Turkish LLMs and their English counterparts.

Compared to these existing benchmarks, the Turkish MMLU (TR-MMLU) benchmark offers several key advancements:
- TR-MMLU is natively designed for Turkish, avoiding the translation-related errors and cultural mismatches often present in multilingual benchmarks.
- The dataset encompasses a wide range of topics, including healthcare, law, history, and the natural sciences, providing a holistic evaluation framework that reflects the linguistic and conceptual diversity of Turkish.
- TR-MMLU evaluates both instruction-following capabilities and knowledge comprehension, enabling a comprehensive assessment of LLM performance in Turkish.

In summary, while existing benchmarks such as TurkishMMLU, OpenLLMTurkishLeaderboard, and THQuAD have laid the groundwork for Turkish NLP, they leave critical gaps in terms of linguistic depth and evaluation scope. TR-MMLU addresses these gaps by providing a robust, transparent, and culturally relevant framework tailored to Turkish. By fostering advancements in model evaluation and development, TR-MMLU sets a new standard for future research and innovation in Turkish NLP.
\section{Task Definition and Methodology}

The primary objective of this study is to establish a robust and comprehensive benchmark for evaluating the performance of Large Language Models (LLMs) in the Turkish language. The Turkish MMLU benchmark (TR-MMLU) was developed using a high-quality dataset of 6,200 multiple-choice questions spanning 62 categories. These categories encompass diverse fields such as law, healthcare, history, and the arts, with questions carefully derived from the Turkish education system and other specialized domains. The benchmark aims to provide a standardized and transparent framework to evaluate the linguistic and conceptual understanding of LLMs, addressing critical gaps in Turkish NLP research.

TR-MMLU serves three core purposes. First, it evaluates Turkish language performance by objectively measuring LLMs' comprehension across diverse topics, offering insights into their strengths and weaknesses. Second, it ensures transparency and reproducibility by making all questions, answers, and evaluation scripts publicly available, allowing independent verification of findings. Third, it contributes to the Turkish AI ecosystem by identifying gaps in model performance and guiding the development of accurate and robust Turkish language models.

The importance of TR-MMLU lies in its ability to address specific challenges faced by Turkish NLP. Unlike existing multilingual benchmarks, which often rely on translations, TR-MMLU is natively crafted by curriculum experts in Turkish, ensuring cultural and linguistic relevance while eliminating errors introduced by translation. Moreover, the dataset is curated to avoid overlap with pre-training data, enabling unbiased evaluations of model performance. This makes TR-MMLU particularly effective for assessing models’ ability to process Turkish text independently of prior exposure.

The methodology for TR-MMLU evaluation is designed to ensure consistency and accuracy. A total of 39 LLMs were evaluated, encompassing open-source models (e.g., Llama, Gemma) and closed-source models (e.g., GPT-4, Claude). The evaluation process includes controlled hardware environments to ensure comparable results, with metrics such as accuracy, success rate, and processing time used to assess performance. To address response format variations, paraphrase detection models are employed to standardize responses and ensure semantic consistency.

Prompt engineering plays a critical role in maximizing the accuracy of model responses. Various prompt structures were tested, each designed to align with the linguistic characteristics of Turkish and the capabilities of the evaluated models. Results were analyzed based on the number of correct answers provided under different prompt conditions, offering insights into the effectiveness of prompting strategies.

A publicly accessible leaderboard has been created on the Hugging Face platform, ranking models based on their accuracy and performance metrics. The leaderboard includes detailed information about each model, such as architecture, parameter size, and quantization level, providing a transparent view of model capabilities. This ensures that the research community has a reliable reference for comparing model performance and identifying best practices for Turkish NLP.

TR-MMLU represents a significant advancement in Turkish NLP by providing a reliable benchmark tailored to the unique characteristics of the Turkish language. This benchmark sets a high standard for evaluating LLMs, offering both researchers and developers a valuable tool to advance the field. By fostering transparency, reproducibility, and collaboration, TR-MMLU lays the groundwork for future innovations in NLP, particularly for resource-limited languages like Turkish.
\section{Experiments and Results}

Experiments were conducted on the Ollama platform using a Python script that automated the evaluation of 39 large language models (LLMs) on the 6,200 questions in the TR-MMLU dataset. To ensure reproducibility, all tests were conducted with a fixed seed value of 42. Various prompt structures tailored to Turkish were employed to maximize response accuracy. Below are some example prompts used during the evaluation:

\textbf{Prompt Examples:}
\begin{itemize}
    \item \textbf{Prompt 1:} \textit{Sana soru ve seçenekleri veriyorum. Sadece hangi seçeneğin sorunun doğru cevabı olduğunu yaz.}  
    \textbf{(Translation:} I will give you a question and its options. Write only the correct answer.\textbf{)}  
    Results: \texttt{gemma2:9b = 63 correct, llama3.1 = 47 correct}
    
    \item \textbf{Prompt 2:} \textit{Sana çoktan seçmeli soru ve seçeneklerini veriyorum. Sorunun doğru seçeneğini bul ve sadece doğru seçeneğin hangi şıkka ait olduğunu söyle.}  
    \textbf{(Translation:} I will give you a multiple-choice question. Indicate which option corresponds to the correct answer.\textbf{)}  
    Results: \texttt{gemma2:9b = 60 correct, llama3.1 = 33 correct}
    
    \item \textbf{Prompt 3:} \textit{Sana soru ve seçenekleri veriyorum, sorunun cevabının hangi seçenek olduğunu bul ve sadece doğru seçeneğin hangi şıkka ait olduğunu söyle.}  
    \textbf{(Translation:} I will give you a question. Find and indicate the correct option.\textbf{)}  
    Results: \texttt{gemma2:9b = 58 correct, llama3.1 = 37 correct}

    \item \textbf{Prompt 4:} \textit{Sana vereceğim çoktan seçmeli sorunun sadece doğru şıkkının harfini söyle.}  
    \textbf{(Translation:} For the question, write only the letter of the correct answer.\textbf{)}  
    Results: \texttt{gemma2:9b = 55 correct, llama3.1 = 36 correct}
    
    \item \textbf{Prompt 5:} \textit{Sana soru ve seçenekleri veriyorum. Sadece hangi seçeneğin doğru olduğunu yaz. Örneğin 'A' veya 'B'. Lütfen açıklama yapma.}  
    \textbf{(Translation:} I will give you a question and options. Write only the correct option, e.g., 'A' or 'B,' without explanation.\textbf{)}  
    Results: \texttt{gemma2:9b = 58 correct, llama3.1 = 42 correct}
\end{itemize}

The evaluation results were published on the Hugging Face platform as three distinct datasets:
\begin{enumerate}
    \item \textbf{AI Turkish MMLU Leaderboard:} This dataset ranks models based on overall performance, including metrics such as accuracy, parameter size, quantization level, and processing time. Table~\ref{tab:leaderboard} summarizes the performance of selected models.
    \item \textbf{AI Turkish MMLU Section Results:} This dataset provides detailed performance analyses across 62 categories, highlighting strengths and weaknesses. Table~\ref{tab:section_results} illustrates the performance of select models in key fields such as the Medical Specialization Examination (TUS) and Public Personnel Selection Exam (KPSS).
    \item \textbf{AI Turkish MMLU Model Responses:} This dataset includes detailed answers from all models, allowing error analysis and deeper insights into model behavior.
\end{enumerate}

\begin{table}[h]
\centering
\caption{AI Turkish MMLU Leaderboard}
\label{tab:leaderboard}
\resizebox{\textwidth}{!}{%
\begin{tabular}{|l|l|l|l|l|l|l|}
\hline
\textbf{Model} & \textbf{Family} & \textbf{Param Size} & \textbf{Quantization} & \textbf{Correct Answers} & \textbf{Accuracy (\%)} & \textbf{Time (s)} \\ \hline
gpt-4o & GPT & Unknown & None & 5260 & 84.84 & 5021 \\ \hline
claude-3.5 & Sonnet & Unknown & None & 5233 & 84.40 & 7379 \\ \hline
llama3.3:latest & llama & 70.6B & Q4\_K\_M & 4924 & 79.42 & 13355 \\ \hline
gemini-1.5-pro & Gemini & Unknown & None & 4758 & 76.74 & 4985 \\ \hline
gemma2:27b & gemma2 & 27.2B & Q4\_0 & 4470 & 72.10 & 5506 \\ \hline
\end{tabular}%
}
\end{table}

\begin{table}[h]
\centering
\caption{AI Turkish MMLU Section Results (Selected Categories)}
\label{tab:section_results}
\resizebox{\textwidth}{!}{%
\begin{tabular}{|l|l|l|l|l|l|}
\hline
\textbf{Model} & \textbf{General Avg (\%)} & \textbf{TUS (\%)} & \textbf{KPSS (\%)} & \textbf{Driver’s License (\%)} & \textbf{AÖF Avg (\%)} \\ \hline
gpt-4o & 84.84 & 91 & 74.5 & 97 & 84.55 \\ \hline
claude-3.5 & 84.40 & 88 & 71.5 & 96 & 84.65 \\ \hline
llama3.3:latest & 79.42 & 85 & 66.5 & 92 & 79.58 \\ \hline
gemma2:27b & 72.10 & 77 & 60 & 90 & 72.57 \\ \hline
aya-expanse:32b & 70.66 & 69 & 55.5 & 84 & 70.96 \\ \hline
\end{tabular}%
}
\end{table}

To address variations in response format, a semantic similarity model, “paraphrase-multilingual-mpnet-base-v2,” was employed. This model computes similarity scores between the generated responses and correct answers. The highest-scoring option was selected as correct if it semantically aligned with the ground truth.

The results revealed that models with robust tokenization strategies tailored to Turkish morphology consistently outperformed others. Fine-tuned models achieved substantial performance gains, though challenges such as \textit{catastrophic forgetting} were observed, where previously acquired knowledge deteriorates during fine-tuning. These findings underscore the importance of localized datasets and optimized training strategies.

Expanding TR-MMLU beyond multiple-choice questions to include open-ended tasks, sentiment analysis, and named entity recognition would further enhance its utility. Additionally, the development of diverse, high-quality Turkish datasets remains a critical priority for advancing Turkish NLP.
\section{Conclusion}

The evaluation of Turkish large language models (LLMs) using the TR-MMLU benchmark has revealed several challenges and opportunities for advancing Turkish natural language processing (NLP). One of the most significant challenges is tokenization, particularly given Turkish’s agglutinative structure and complex morphology. Many existing models struggle to effectively process Turkish tokens, leading to suboptimal accuracy. Future work should focus on developing advanced tokenization techniques tailored to Turkish, addressing linguistic complexities and enabling more accurate model representations.

Fine-tuning strategies for Turkish-specific tasks present another area for improvement. While fine-tuning on Turkish datasets has proven to enhance performance, it is often accompanied by challenges such as catastrophic forgetting, where previously learned knowledge degrades during the fine-tuning process. To mitigate this, future research should explore more robust fine-tuning methodologies that preserve foundational knowledge while optimizing models for Turkish-specific tasks.

The limited availability of high-quality Turkish datasets is another barrier to comprehensive model evaluation and training. Expanding the pool of diverse and ethically sourced Turkish datasets will not only support fine-tuning but also improve generalizability across various NLP applications. Additionally, broadening the TR-MMLU benchmark to include open-ended tasks, sentiment analysis, named entity recognition, and contextual word embeddings would enable a more holistic evaluation of Turkish language models and their capabilities.

Model performance discrepancies observed across different categories also highlight the need for deeper investigations into the underlying causes of these variations. Understanding these discrepancies will clarify the strengths and limitations of Turkish LLMs and provide guidance for targeted improvements. Such research will enable models to better adapt to the linguistic nuances of Turkish, enhancing their versatility across different applications.

TR-MMLU represents a pivotal step forward for Turkish NLP, providing a robust and transparent evaluation framework designed specifically for the Turkish language. By offering 6,200 multiple-choice questions derived from the Turkish education system and specialized fields, this benchmark serves as an invaluable resource for researchers and developers to objectively measure model performance. 

The findings from this study underscore the critical importance of tokenization and language-specific fine-tuning. Models equipped with advanced tokenization techniques tailored to Turkish morphology consistently achieved higher accuracy. Additionally, fine-tuning with Turkish-specific data yielded significant performance gains, though with some limitations in maintaining foundational knowledge. These results highlight the need for tailored approaches in model design and training to address the unique challenges of Turkish NLP.

As a reliable benchmark emphasizing transparency and reproducibility, TR-MMLU fosters collaboration and innovation within the research community. Moving forward, the benchmark aims to expand its scope beyond multiple-choice questions, incorporating more diverse evaluation tasks to provide a comprehensive assessment of Turkish LLMs. This will not only enhance the utility of TR-MMLU but also set new standards for evaluating language models in resource-limited languages.

In conclusion, TR-MMLU is a foundational advancement for Turkish NLP, offering a standardized framework for evaluating language models and guiding future research. By addressing key challenges such as tokenization, fine-tuning, and dataset availability, and by expanding evaluation metrics, researchers and developers are encouraged to enhance Turkish language processing technologies further. These advancements will undoubtedly contribute to setting new benchmarks in the field and drive the development of more robust and versatile Turkish LLMs.

\bibliographystyle{unsrt}
\bibliography{references}

\end{document}